\documentclass[a4paper,twoside]{article}

\usepackage{epsfig}
\usepackage{subcaption}
\usepackage{calc}
\usepackage{amssymb}
\usepackage{amstext}
\usepackage{amsmath}
\usepackage{amsthm}
\usepackage{multicol}
\usepackage{pslatex}
\usepackage{apalike}
\usepackage[bottom]{footmisc}
\usepackage{SCITEPRESS}     

\begin{document}

\title{Concept explainability for plant diseases classification}

\author{\authorname{Jihen Amara\sup{1}, Birgitta König-Ries\sup{1,2}\ and Sheeba Samuel\sup{1,2}}
\affiliation{\sup{1}Heinz Nixdorf Chair for Distributed Information Systems, Department of Mathematics and Computer Science, Friedrich
Schiller University Jena, Jena, Germany
}
\affiliation{\sup{2}Michael-Stifel-Center for Data-Driven and Simulation Science, Jena, Germany}
\email{\{jihene.amara, birgitta.koenig-ries, sheeba.samuel\}@uni-jena.de }
}


\abstract{Plant diseases remain a considerable threat to food security and agricultural sustainability. Rapid and early identification of these diseases has become a significant concern motivating several studies to rely on the increasing global digitalization and the recent advances in computer vision based on deep learning. In fact, plant disease classification based on deep convolutional neural networks has shown impressive performance.
However, these methods have yet to be adopted globally due to concerns regarding their robustness, transparency, and the lack of explainability compared with their human experts counterparts. Methods such as saliency-based approaches associating the network output to perturbations of the input pixels have been proposed to give insights into these algorithms. Still, they are not easily comprehensible and not intuitive for human users and are threatened by bias. In this work, we deploy a method called Testing with Concept Activation Vectors (TCAV) that shifts the focus from pixels to user-defined concepts. To the best of our knowledge, our paper is the first to employ this method in the field of plant disease classification. Important concepts such as color, texture and disease related concepts were analyzed. The results suggest that concept-based explanation methods can significantly benefit automated plant disease identification.
}

\onecolumn \maketitle \normalsize \setcounter{footnote}{0} \vfill

\section{\uppercase{Introduction}}
\label{sec:introduction}

Plant diseases are important factors as they result in serious reduction in quality and quantity of agricultural products. Therefore, early detection and diagnosis of these diseases are important. In our prior work, we built a deep learning model based on convolutional neural networks (CNN) to identify diseases from images of plant leaves \cite{amara2017deep} automatically. While successful, this type of model is a black-box predictor preventing the acquisition of any explanation for the predictions.
We believe that the availability of an explainable model that can rapidly and accurately identify and quantify plant diseases would have a significant impact on scientific research and smart crop production.  Humans need to know and understand about the detection, symptoms and diagnosis process in addition to the high accuracy of the plant disease classification models.
Hence, it is widely believed that coupling black-box models with interpretability techniques would increase their adoption in the industry, agriculture, healthcare, and other high-stakes fields \cite{molnar2020interpretable}. In this paper, we define interpretability as the ability to explain or present in understandable terms to a human, as suggested by \cite{doshi2017towards}. We will use explainability and interpretability interchangeably in the paper.
This urgent need for model interpretability led to a proliferation of proposed methods. These methods, that we will review in Section 2, follow a common strategy which is simply highlighting pixels that were relevant for a certain class classification by a neural network. However, they suffer from various drawbacks. It has been shown that they are not as reliable as expected and are susceptible to human confirmation biases \cite{ghorbani2019interpretation}. 
Consequently, a new line of research has focused on producing explanations in the form of high-level "concepts" \cite{kim2018interpretability}. 
Hence, our goal in this work is to investigate the usefulness of these methods that focus on producing semantic and human-understandable explanations for our use case plant diseases classification. We believe that semantics can enhance interpretability in many areas. Instead of just displaying numbers or saliency maps on image regions, these methods output explanations that are understandable by humans and based on interpretable concepts.
For a diseased plant leaf, instead of outputting a single probability (e.g., 90\% probability of having the late blight disease), this type of algorithm would, for example, output "high texture irregularity, high amount of black, yellow and brown areas on top of the leaf" etc.
One of the methods that belong to this family is Testing with Concept Activation Vectors (TCAV) \cite{kim2018interpretability}. This method  tests the sensitivity of a trained deep neural model to a defined concept of interest. It also provides a global explanation for the model. The central idea of TCAV is to evaluate how responsive a CNN is to input patterns representing a concept (e.g., color or texture) linked to the prediction output of the CNN (e.g., the class "late blight" disease). 
Therefore, in our work, interpretability refers to a quantitative explanation of which plant disease concepts are most important for accurate plant disease classification by CNNs.
Our research contributions are summarized as follows. First, to the best of our knowledge, this study is the first attempt at a comprehensive understanding of what semantic concepts the CNN learns during the plant disease diagnoses. This is a critical issue for the vast proliferation of deep learning techniques in plant phenotyping tasks. It can give insights about CNN models for plant image analysis and help increase trust in such models. Second, we have presented a concept dataset that can be used in the future to test concept methods with plant disease image classification.
The remainder of this paper is organized as follows. Section 2 presents related work on interpretability and its application to plant disease classification, while Section 3 describes the trained networks, the TCAV theory, and the utilized datasets. Section 4 presents our experimental results regarding prediction accuracy and model interpretation. Finally, Section 5 depicts our conclusive remarks and possible future work.
\section{\uppercase{Related Work}}

In recent years, there has been an increasing interest in explainability and interpretability approaches to deep learning. 
Two main sets of methods have been proposed, which are saliency-based methods \cite{zeiler2014visualizing} and concept-based methods \cite{kim2018interpretability,zhou2018interpretable}. 
The first is based on simply highlighting relevant pixels for a certain class classification by a neural network. Saliency-based methods are also called feature attribution methods. In the case of image classification, features are input pixels, and such methods aim to give each pixel a value that can be understood as the pixel's relevance to the image's classification.
Few works in the literature focus on interpreting deep learning models for plant disease classification using saliency-based methods. For example, different papers have tried to apply these visualization methods and present a comparison study when applied to plant diseases \cite{brahimi2018deep,toda2019convolutional,kinger2021explainable}. Other works have focused on using visualization methods to extract the description of plant diseases from trained CNN \cite{sladojevic2016deep,ballester2017assessing}.
In addition, some papers tried to present novel visualization methods for plant disease classification \cite{ghosal2018explainable,brahimi2019deep}. These methods are beneficial because they give visual explanations, making it easy to see the critically highlighted pixels. However, these methods are considered fragile and sensitive to adversarial perturbation \cite{ghorbani2019interpretation}. Other work has shown how these methods could be highly unreliable \cite{kindermans2019reliability}.
Since these methods create importance maps based on individual input samples, they provide only local interpretations and cannot explain the network's decisions on a global scale \cite{lucieri2020interpretability}.
In addition, these methods' lack of expressiveness to users is an essential drawback. For instance, the importance of a single pixel in the classification does not bring a meaningful explanation, and it is also contrived by the number of features \cite{molnar2020interpretable}. 
Hence, concept-based approaches were proposed to address these limitations \cite{zhou2018interpretable,kim2018interpretability}. Concepts can be colors, objects, or abstract ideas. Users can define these concepts without the need to train the network on them. 
Hence, they are understandable by humans, and they are not limited to the neural network feature space. 
One of these methods is the TCAV approach, which was proposed by \cite{kim2018interpretability}. We will describe this method in detail in the following section. 
TCAV was successfully used with a different application in the medical field \cite{lucieri2020interpretability}. To our knowledge, concept-based explanation methods have not previously been explored for plant disease classification networks.
Hence, in this work, we adopt the TCAV method to the problem of plant disease classification. We also present a set of concepts that could be used in this case.

\section{\uppercase{Materials and methods}}
\subsection{Dataset and networks for plant diseases classification}
\subsubsection{Datasets}
The Plant Village dataset is a public repository that contains 54,323 images of 14 crops and 38 different types of plant diseases \cite{hughes2015open}. It has been extensively used by the community of plant disease image classification. We used only images of tomato leaves from this dataset to train the CNNs. The total number of images is 18,160, divided into ten classes (nine diseases and a healthy class). Finally, the data was separated into three sets, containing 80\% of the data in the training set, and the remaining 20\% is divided equally between the testing and validation sets. 
Figure 1 presents one example of each disease class, and Table 1 summarizes our dataset.

\begin{figure}[!h]
  \centering
   {\epsfig{file = 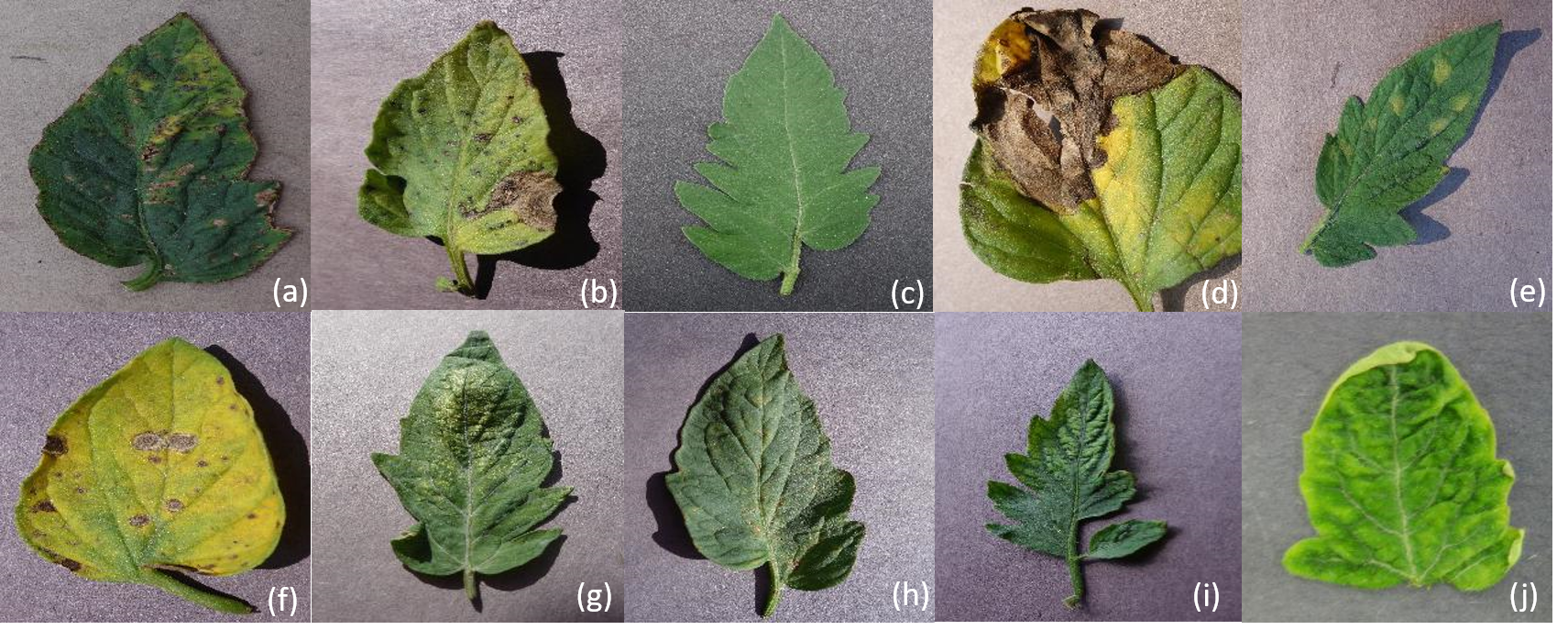, width = 7.5cm}}
  \caption{Sample images from the Plant Village Dataset. (a) Bacterial Spot, (b) Early Blight,  (c) Healthy, (d) Late Blight, (e) Leaf Mold, (f) Septoria Leaf Spot, (g) Spider Mites, (h) Target Spot, (i) Mosaic Virus, and (j) Yellow Leaf Curl Virus.
  }
  \label{fig:example1}
 \end{figure}

\begin{table}[h]
\caption{The image numbers of each tomato disease class in the Plant Village dataset.}\label{tab:example1} \centering
\begin{tabular}{|c|c|}
  \hline
  Classes & Number of Images \\
  \hline
  Bacterial Spot & 2127 \\
  \hline
  Early Blight & 1000 \\
  \hline
  Healthy & 1591 \\
   \hline
     Late Blight & 1909 \\
   \hline
   Leaf Mold & 952 \\
   \hline
    Septoria Leaf Spot & 1771 \\
   \hline
    Spider Mites & 1676 \\
   \hline
    Target Spot & 1404 \\
   \hline
    Mosaic Virus & 373 \\
   \hline
    Yellow Leaf Curl Virus & 5357 \\
   \hline
   Total & 18,160 \\
   \hline
\end{tabular}
\end{table}

\subsubsection{Models}
In this work, we will focus on two widely used CNN architectures: InceptionV3 \cite{szegedy2016rethinking} and Vgg16 \cite{szegedy2015going}. We selected these architectures due to their high use in the plant disease classification literature \cite{lee2020new}. 
To train these networks, we took advantage of fine-tuned transfer learning \cite{weiss2016survey}, which is based on transferring the knowledge gained from training the models on a more significant dataset to a smaller one. The models were created and loaded with pre-trained weights on the ImageNet dataset \cite{deng2009imagenet}. In addition, excluding top layers was performed by defining new layers on the top of the models. The altered top architecture consists of three dense layers with corresponding dropout layers.
For training and optimizing the weights on the tomato disease dataset, we froze the first 51 convolutional layers and made the rest trainable for InceptionV3, and we unfroze the final convolutional layers of Vgg16 and made them trainable. Training optimization was carried out via stochastic gradient descent optimizer with a learning rate of 0.0001 and momentum of 0.9. We used a batch size of 20 and 20 epochs for training. We use data augmentation techniques to increase the dataset size in training while including different variations. These variations consist of transformations such as random rotations, zooms, translations, shears, and flips to the training data as we train.
Both models were implemented using Keras \cite{chollet2021deep}, and were saved for subsequent performance testing and interpretability analysis. We experimented on a server with a GPU that consists of two NVIDIA Tesla V100 with 128 GB of RAM. 

\subsection{Network explanation through concepts}
This section will introduce a deep neural network explanation method called Testing with CAV (TCAV). TCAV requires three types of data samples: samples representing particular plant disease concepts (concept samples), samples known to be specific plant disease classes (disease-class samples), and samples selected randomly to supervise the training quality (to ensure the stability of the results). In contrast to the original work \cite{kim2018interpretability}, where random examples may by chance include the concept, we ensure that the selected set is strictly negative. Below is a detailed description of TCAV theory and our study's used concepts.
\subsubsection{TCAV theory}
To better understand the concepts used by the trained CNN to classify images as either healthy or diseased leaves, we use the concept activation vectors (CAVs) method \cite{kim2018interpretability} defined as follows:
Importance of a ‘concept’ C (e.g., brown)  to an image class $k$ (e.g., late blight) is found by taking the directional derivative of class predictions (for class $k$) at each layer $l$ of a CNN in the direction of (with respect to) a CAV. 
 Hence, a CAV, is a vector ${v_{c}^{l}}$ in the embedding space of a CNN denoting the direction that encodes the given concept C in the activation space of a neural network layer $l$.
To find CAV of a concept C, we need to prepare two datasets: a concept dataset, which represents C, and a random dataset that does not contain the concept. 
A binary concept classifier is trained to separate the activations generated by the concept set from those generated by the random set at a specific hidden layer $l$.
The CAV ${v_{c}^{l}}$ is then defined as the normal to the hyperplane separating the two classes at a particular layer $l$.
Finally, given an image input $x$, we can measure its conceptual sensitivity by computing the directional derivative S of the prediction in the direction of the CAV ${v_{c}^{l}}$ for concept C:
\begin{equation}\label{eq1}
\begin{split}
   \displaystyle  S_{c,k,l}(x) = \lim_{\epsilon  \to 0} \frac{h_{l,k}(f_{l}(x)+\epsilon v_{c}^{l})-h_{l,k}(f_{l}(x))}{\epsilon} \\
    = \nabla h_{l,k}(f_{l}(x)). v_{c}^{l}\\
    \end{split}
\end{equation}

where $f_{l}$ maps the input $x$ to the activation vector of the layer $l$  and $h_{l,k}$ maps the activation vector to the logit output of class $k$.
Then, to measure the influence of a CAV on a class of input images, a metric called TCAV score is computed. It employs the directional derivatives $S_{c,k,l}(x)$ to compute the contextual sensitivity of a concept towards the whole inputs $X_{k}$ for class $k$.
The TCAV score is given by:
\begin{equation}\label{eq2}
TCAV_{Q_{c,k,l}}=\frac{\left|x\in X_{k} ; S_{c,k,l}(x)> 0 \right|}{\left| X_{k} \right|}
\end{equation}
Hence, the TCAV score denotes the ratio of class $k$’s inputs that are positively influenced by concept $C$.
Additionally, the authors perform a statistical significance test of  TCAV scores to make sure that only meaningful $CAVs$ are taken into account. 
They compute multiple CAVs between concept images and a batch of random images. In addition, they train random CAVs where both concept set and random set are random images. Then they perform a two-sided t-test of the TCAV scores based on these multiple samples.  The resulting concept is considered significant for the class prediction if they can reject the null hypothesis of a TCAV score of 0.5. This helps to make sure that concept CAVs and random CAVs are significantly different from each other. TCAV authors further perform Bonferroni correction $( p<\alpha /m,  m=2)$ for multiple comparisons between all concept-random pairs to reduce potential for false positives (incorrect rejection
of the null hypothesis, or a Type I error) to prevent mistaking as significant a truly insignificant concept \cite{kim2018interpretability}.
In our work,  TCAV  is implemented using Keras rather than the original authors' Tensorflow code \cite{kim2018interpretability}.

\subsubsection{Concepts used for analysis}
Identification of plant diseases is commonly accomplished through visual inspection of the disease's effect on the plant. This effect is considered a symptom and could be detectable as a change of color or texture of the leaf caused by the pathogen. In our work, we would like to see if the CNN learned representation of various concepts and those often used by plant diseases experts are complementary. Hence, we choose to test the concepts of color, texture, and late blight disease pattern described below. Also, to show the feasibility of this methodology and to have a general disease pattern, we focus on one disease class: late blight disease, which affects both potatoes and tomatoes. 
The concepts used in this work to interpret the deep classifiers are defined below in accordance with the description in plant stress phenotyping literature \cite{isleib2012signs}.

  \noindent {\bf Color concepts:}  
To train TCAV, we used colors as concepts such as red, brown, blue, yellow, and green. We wanted to test our trained models' sensitivity to these colors especially green, brown, and yellow which can present the difference between healthy and diseased states. We will hence gain more comprehension of how our internal network behaves. Random images used are grayscale images of diseased plant leaves  other than tomato. We produced 100 pictures per color synthetically by generating the color channel randomly.

   \noindent {\bf Describable Texture dataset (DTD) concepts  \cite{cimpoi2014describing}:}  
Texture is an essential part of plant disease identification as it depicts more knowledge of the infected leaf region. Infected leaves can turn dry and present signs like crackedness, wrinkles, bumpiness, fibrousness, etc. Hence, this work presents a dataset of texture concepts that could be used with plant diseases and are also understandable to humans.
The dataset is extracted from the DTD database \cite{cimpoi2014describing}. DTD is a texture database collected “in the wild” with semantic attributes. It contains a list of 47 categories inspired by human perception. The concepts concerning plant diseases that we choose from DTD are: blotchy, bumpy, cracked, fibrous, pitted and wrinkled. For the TCAV experiment, we chose 100 images for each concept category. For the random images, we selected healthy images of leaves other than Tomato.
Figure 2 presents some examples from the DTD categories.
\begin{figure}[!h]
  \centering
   {\epsfig{file = 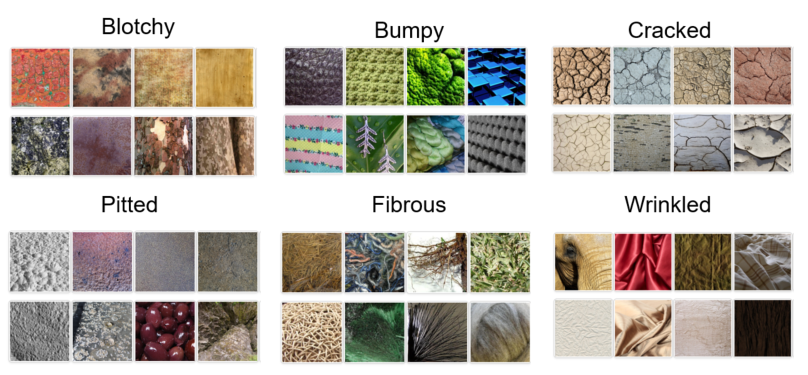, width = 7.5cm}}
  \caption{Concepts related to plant disease symptoms extracted from DTD texture database (blotchy, bumpy, cracked, pitted, fibrous, wrinkled) \cite{cimpoi2014describing}.}
  \label{fig:example1}
 \end{figure}
 
\noindent {\bf Late Blight disease concepts:}
Late blight is a devastating disease that can affect potato and tomato plants. It is caused by the water mold Phytophthora infestans \cite{britannica2020editors}. The first symptoms of the disease can start with light to dark green spots in the leaves, which rapidly develop into large dark brown or black lesions. Leaf lesions are also frequently surrounded by a yellow chlorotic halo. Leaves can become dry and shriveled, and crops can be severely damaged. In our experiment, we use the images of potato leaves infected by the late blight disease as a visual representation of the disease concept. These images will depict the disease pattern that appears on the infected leaves. It will help us test the network's ability to learn the representation of the disease independently from the leaf shape or type. We used healthy leaves of plants other than tomatoes and potatoes for the random images.

\section{\uppercase{Results and analyzes}}
\subsection{Network training results}

This section will present the results of the models evaluation on our dataset. Table 2  shows the found results. The performance of the trained models is evaluated using recall, precision, and accuracy metrics \cite{davis2006relationship}.

\begin{table}[h]
\caption{Performance measures for pre-trained models Vgg16 and InceptionV3.}\label{tab:example2} \centering
\begin{tabular}{|c|c|c|}
  \hline
   & InceptionV3 & Vgg16 \\
  \hline
  Accuracy  & 0.92 & 0.98 \\
  \hline
  Precision  & 0.93 & 0.98 \\
  \hline
   Recall  & 0.92 & 0.98 \\
  \hline
   F1-score  & 0.92 & 0.98 \\
  \hline
\end{tabular}
\end{table}
As shown in Table 2, Vgg16 and InceptionV3 trained on the Tomato disease dataset have achieved good results with a predominance from Vgg16 achieving an accuracy of 98\% compared to InceptionV3 with an accuracy of 92\%.
\subsection{TCAV experiments Result}
We will concentrate on the quantitative evaluation of the TCAV score. The TCAV score quantifies a given concept's positive or negative influence on a specific target class. For InceptionV3, in each experiment, TCAV scores were computed within 9 layers (mixed3, mixed5 to mixed9), which are the concatenation layers at the end of each inception module. The numbers denotes the location depth of a layer in the network. This can help us see how different network depths affected the final classification. 
For Vgg16, TCAV scores were computed within three layers (flatten, dense, and dense\_1). We tested the final layers added on top of the model, which go from the shallowest (flatten) to the deepest (dense\_1).
We train CAVs on the activations extracted from these chosen layers.

\subsubsection{Importance of Colors concepts}

\begin{figure}[!h]
  \centering
   {\epsfig{file = 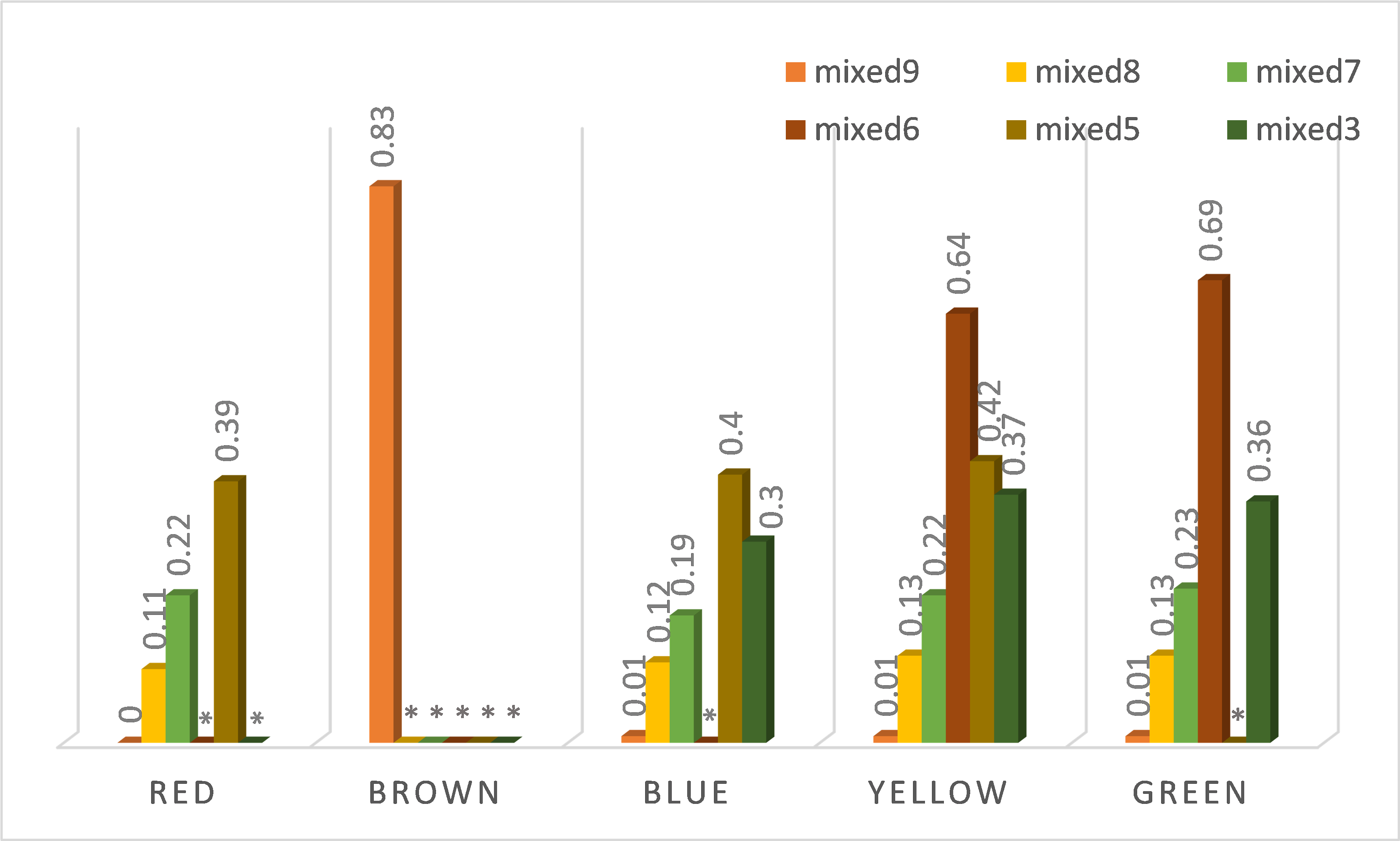, width = 7.5cm}}
  \caption{TCAV scores for color concepts red, brown, blue, yellow and green in InceptionV3 with layers mixed 3, mixed 5, mixed 6, mixed 7, mixed 8 and mixed 9.}
  \label{fig:tcavcolor}
 \end{figure}
 
 \begin{figure}[!h]
  \centering
   {\epsfig{file = 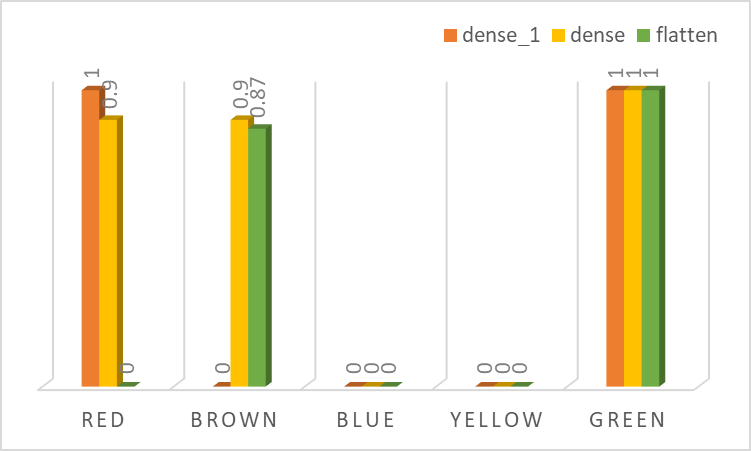, width = 7.5cm}}
  \caption{TCAV scores for color concepts red, brown, blue, yellow and green in Vgg16 with layers dense\_1, dense and flatten.}
  \label{fig:vggcolor}
 \end{figure}

Results for InceptionV3 are shown in Figure \ref{fig:tcavcolor}, where we tested layers  mixed 3, mixed 5, mixed 6, mixed 7, mixed 8, and mixed 9. The layers that should be chosen for testing are those who passed the statistical testing (t-test) mentioned above successfully for most of the concepts.
 In consistency with plant diseases experts description \cite{isleib2012signs} we found that brown and yellow were both significant for identifying late blight disease, with TCAV scores as high as 0.83 for brown in mixed 9 and 0.64 for yellow in mixed 6. Also, green was significant in layer mixed 6 with a TCAV score of 0.69. In contrast, the rest of the colors were less important. The bars marked by a star ‘*’ indicate TCAV scores that did not pass the statistical testing.
 For Vgg16 (Figure \ref{fig:vggcolor}), the important color concepts captured by the model were green and brown. To our surprise, it seems red was also crucial for the model in classifying the disease. Green had a TCAV score of 1 for the three tested layers dense\_1, dense, and flatten. While brown had TCAV scores of 0.9 and 0.87 for the dense and flatten layers. Moreover, red had TCAV scores of 1 and 0.9 for layers dense\_1 and dense.
Hence, even though Vgg16 had better classification performance than InceptionV3, the latter was better at capturing color concepts similar to those used by experts to classify plant diseases.
\subsubsection{Importance of Describable Texture dataset (DTD) concepts}
To show our trained models' sensitivity to textures related to plant disease symptoms, we experiment on a texture recognition dataset (DTD).
\begin{figure}[!h]
  \centering
   {\epsfig{file = 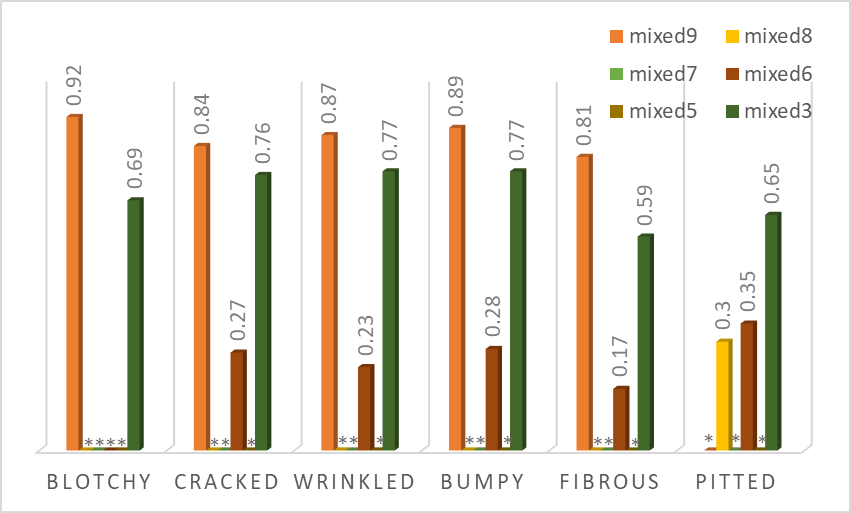, width = 7.5cm}}
  \caption{Conceptual importance (TCAV scores) of DTD concepts blotchy, cracked, wrinkled, bumpy, fibrous, pitted in InceptionV3.}
  \label{fig:inceptiondtd}
 \end{figure}
 \begin{figure}[!h]
  \centering
   {\epsfig{file = 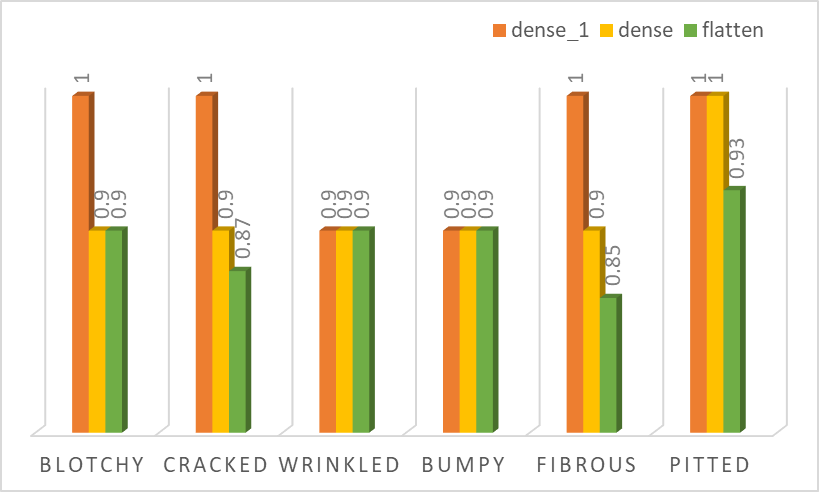, width = 7.5cm}}
  \caption{Conceptual importance (TCAV scores) of DTD concepts blotchy, cracked, wrinkled, bumpy, fibrous, pitted in Vgg16.}
  \label{fig:vgg16dtd}
 \end{figure}
As shown in Figure \ref{fig:inceptiondtd} and Figure \ref{fig:vgg16dtd}, we got a good insight into what texture concepts exactly were important for InceptionV3 and Vgg16 when looking to identify a plant disease. TCAV scores show that concepts such as blotchy, cracked, wrinkled, bumpy, fibrous, and pitted are highly relevant to the late blight disease class.
In the case of InceptionV3 (Figure \ref{fig:inceptiondtd}), we found that the layers that could capture these important texture concepts were mixed 3 and 9, while for the rest of the layers, TCAV scores were insignificant. This finding can raise the possibility that some layers can be omitted and consequently reduce the parameters needed for the network training. Especially in the case of plant disease classification, where mobile phones can be used for such tasks, reducing the number of network parameters is exceptionally beneficial for memory and calculation efficiency.  
In the other case, results show that Vgg16 (Figure \ref{fig:vgg16dtd})  is sensitive to all the DTD concepts compared to InceptionV3. TCAV scores were as high as 1 and 0.9 for the tested layers dense, dense\_1, and flatten, which indicate the sensitivity of these layers to such concepts when classifying late blight disease. 
In the case of our training set, images are about a single leaf in a uniform background, which makes recognizing the change of textures on the leaf sufficient to solve the task of plant disease classification. 
Hence, the results show that different texture detectors emerge, which explains why the texture concepts dominate in the tested layers especially in the case of Vgg16. 
This discovery is consistent with the former qualitative finding from \cite{toda2019convolutional} where feature visualization methods showed that the model focused on learning the visual cues (textutre) of the disease lesions rather than objects or shapes. Hence, TCAV permit for quantitative confirmation of this previous qualitative discovery.
Results look very much as expected and are aligned with the description in plant stress phenotyping literature \cite{isleib2012signs}.
\subsubsection{Importance of late blight disease concepts}

\begin{figure}[!h]
  \centering
   {\epsfig{file = 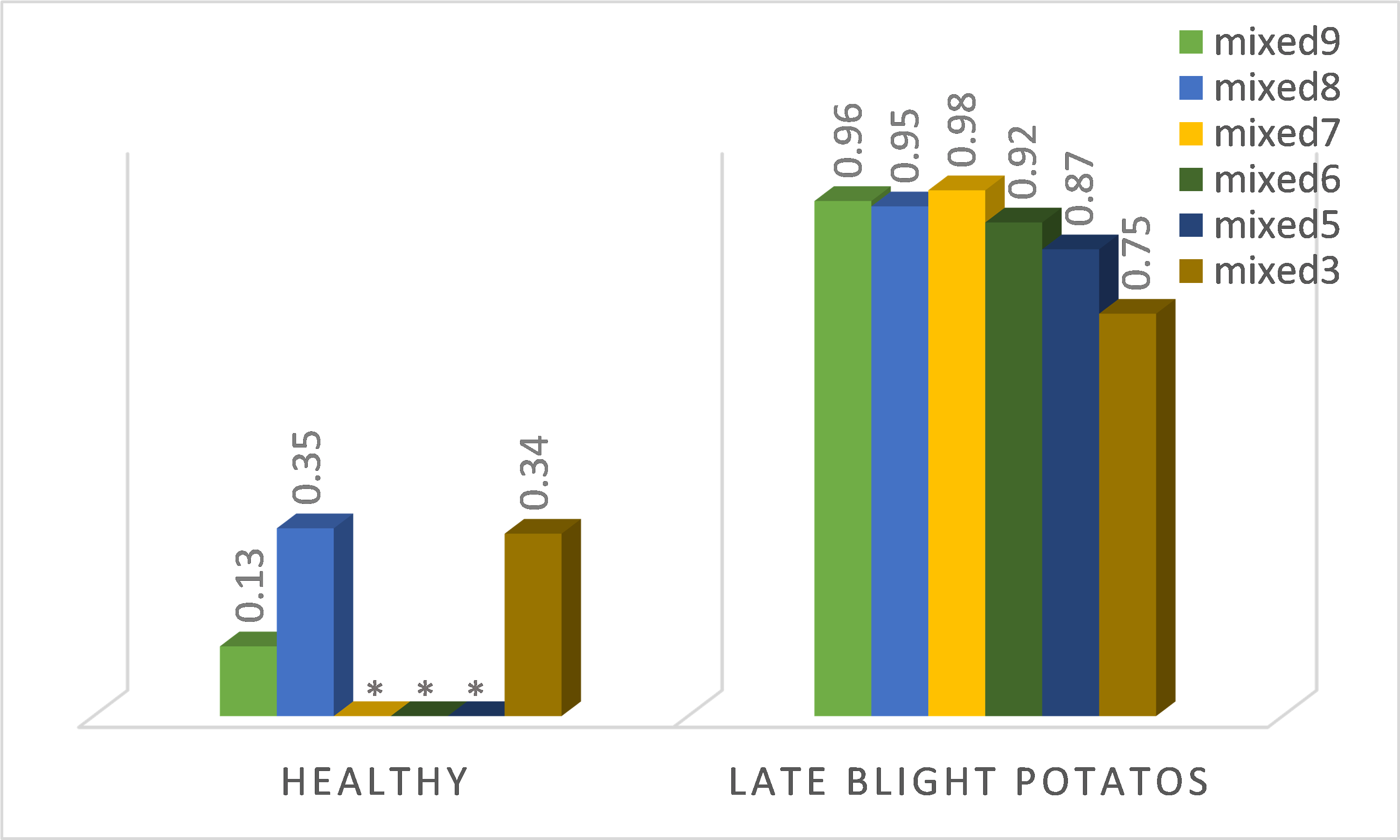, width = 7.4cm}}
  \caption{Conceptual importance (TCAV scores) of late blight disease concept in InceptionV3.}
  \label{fig:incep}
 \end{figure}
 \begin{figure}[!h]
  \centering
   {\epsfig{file = 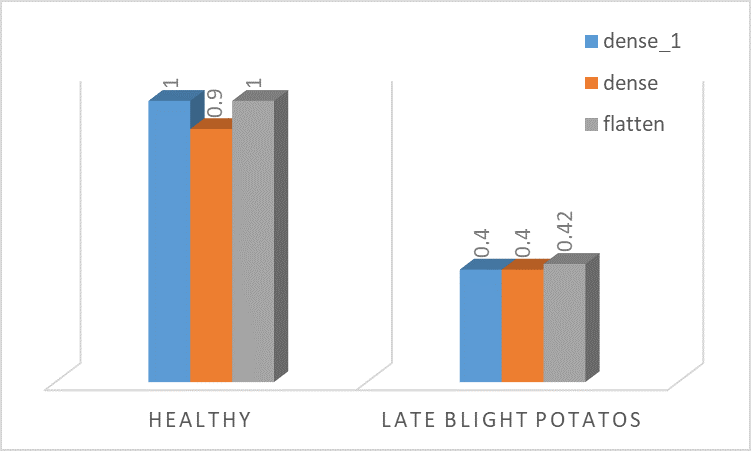, width = 7.4cm}}
  \caption{Conceptual importance (TCAV scores) of late blight disease concept in Vgg16.}
  \label{fig:vgg}
 \end{figure}
To assess the importance of the late blight concept for both InceptionV3 and Vgg16 when classifying the disease, we experimented with images of potato with late blight disease and healthy images of tomato. 
Interestingly, from Figure \ref{fig:incep} we can see that, for the InceptionV3, the late blight concepts contribute positively to the tomato late blight disease class. While the healthy concepts don't have any significant contribution. To our surprise, this was the opposite for Vgg16 (Figure \ref{fig:vgg}) where the late blight concepts did not play a role.
A possible explanation for these results may be the fact that the network did not focus on disease regions but crop-specific characteristics such as leaf shape and also for the fact that InceptionV3 is a deeper network than Vgg16 which helped in capturing better the concepts.
These findings further support the idea of training models with common diseases regardless of crop type for more generalizability \cite{lee2020new}.
Also, it shows that regardless of the accuracy that indicates a good learning performance, DL models could inherently learn or fail to learn representations from the data which an expert might consider important \cite{isleib2012signs}. This enforces the importance of including explainability in the deep learning workflow.

\section{\uppercase{Conclusions}}
\label{sec:conclusion}

The influence of artificial intelligence in modern agriculture is increasing with the incorporation of new technologies such as machine learning and robotics to boost crop abundance and quality. Plant pathologists are starting to use deep learning techniques to help in disease classification, surveillance, and management. However, they have yet to be adopted globally due to concerns regarding robustness, transparency, and lack of explainability compared with their human experts counterparts. Hence, in our work, we adopted a concept-based explanation method called TCAV for plant disease classification to gain more insights into the model and which concepts affect its decisions. Based on transfer learning techniques, we have trained two famous networks, Vgg16 and InceptionV3. Important concepts such as color, texture, and concept-related disease were analyzed. Our results show that deep learning based models can learn and use similar disease-related concepts for prediction as plant pathologists use. Nevertheless, the study could be extended to grant a more comprehensive interpretation of the TCAV scores for this particular use case. It would be great to exploit more granular labeling of disease concepts and explore more disease classes to draw further insight into the model's classification mechanism and to give better validation of its decisions. Future work could incorporate plant disease datasets from the field with real background, and then, we can use TCAV to ensure the model is unbiased towards the background. Also, finding a way to define concepts automatically could simplify the process by eliminating the necessity for manual annotations and could allow revealing new knowledge for plant disease experts or unexpected biases from the network. We would also like to investigate ways of adding semantics through context and Knowledge Graphs into TCAV.

\section*{\uppercase{Acknowledgements}}

We would like to thank Chiheb Karray for helpful discussions. We thank the German Academic Exchange Service (DAAD) for supporting the work of Jihen Amara and the Carl Zeiss Foundation
for the financial support of the project “A Virtual Werkstatt for Digitization in the Sciences (K3)” within the scope of the program line “Breakthroughs: Exploring Intelligent Systems for Digitization”- explore the basics, use applications”. The computational
experiments were performed on resources of Friedrich Schiller University Jena supported in part by DFG grants INST 275/334-1 FUGG and INST 275/363-1 FUGG.

\bibliographystyle{apalike}
{\small
\bibliography{example}}

\end{document}